\pdfoutput=1
\PassOptionsToPackage{x11names}{xcolor}
\documentclass[11pt]{article}

\usepackage{EMNLP2022}
\usepackage{hyperref}
\usepackage{times}
\usepackage{latexsym}

\usepackage{booktabs}
\usepackage{multirow}
\usepackage{graphicx}
\usepackage{array}
\usepackage{todonotes}
\usepackage{amsmath}
\usepackage{linguex}

\usepackage{soul}
\usepackage{bm}
\sethlcolor{lightgray}
\usepackage{siunitx}
\usepackage{microtype}
\usepackage{tikz}
\usepackage{tikz-dependency}
\usetikzlibrary{decorations.pathreplacing}

\usepackage{times}
\usepackage{latexsym}

\usepackage[T1]{fontenc}

\usepackage[utf8]{inputenc}

\usepackage{microtype}

\usepackage{inconsolata}

\title{Leveraging World Knowledge in Implicit Hate Speech Detection}

\author{Jessica Lin \\
  Department of Linguistics\\
  Georgetown University\\
  \texttt{yl1290@georgetown.edu}\\} 

\begin{document}
\maketitle
\begin{abstract}
\textbf{\textit{Warning}}: \textit{This paper contains content that may be offensive or disturbing.}\\
While much attention has been paid to identifying explicit hate speech, implicit hateful expressions that are disguised in coded or indirect language are pervasive and remain a major challenge for existing hate speech detection systems. This paper presents the first attempt to apply Entity Linking (EL) techniques to both explicit and implicit hate speech detection, where we show that such real world knowledge about entity mentions in a text does help models better detect hate speech, and the benefit of adding it into the model is more pronounced when explicit entity triggers (e.g., rally, KKK) are present. We also discuss cases where real world knowledge does not add value to hate speech detection, which provides more insights into understanding and modeling the subtleties of hate speech.
\end{abstract}

\section{Introduction}

Hate speech on social media facilitates the spread of violence in the real world. For this reason, the detection of hatred content online increasingly gains importance. However, most work in hate speech detection has focused on explicit or overt hate speech, failing to capture the implicit hateful messages in coded or indirect language (e.g., sarcasm or metaphor) that disparage a protected group or individual, or to convey prejudicial and harmful views about them \cite{waseem2017understanding}. Examples \ref{ex1} and \ref{ex2} from \citet{elsherief2021latent} show the two types of hate speech, explicit vs. implicit: 

\ex. \textit{\#jews \& n*ggers destroy and pervert everything they touch \#jewfail \#n*ggerfail} (explicit hate speech) \label{ex1}

\ex. \textit{don't worry, charlottesville was just the beginning. we're growing extremely fast} (implicit hate speech, implied statement: larger white supremacist events will happen) \label{ex2}

As shown in \ref{ex1} and \ref{ex2}, explicit hate speech is direct and uses specific keywords while implicit hate speech does not contain explicit hateful lexicon or phrases and often uses coded or indirect languages to disguise the malicious intent \cite{elsherief2021latent}. \par
Modeling implicit sentiment in hate speech is still in its infancy, and the capacity to acquire background knowledge enhances the correct detection of hate speech by machines \cite{kiritchenko2021confronting,li2021sarcasm}. To be able to understand the implied statement of hate speech, machine learning systems need extratextual information that provides world knowledge associated with natural language concepts. For example, it would be impossible for a reader who does not know what happened in Charlottesville to understand the implicit hateful message in \ref{ex2}. The reader wouldn't be able to understand the implied message ``larger white supremacist events will happen'' without knowing that Charlottesville is a metonym for a white supremacist rally that took place in Charlottesville, Virginia in August, 2017. Conversely, background knowledge that reasons about entity mentions in a text could add value to the detection of implicit hate speech. Incorporating such knowledge ideally should make it easier for the learning model to detect hate speech where it is not apparent from the text.\par
With this motivation, this study applies Entity Linking (EL) to identify entities in tweets, link them to an external knowledge base (KB; Wikipedia in this study), and acquire their Wikipedia descriptions that would be encoded with Sentence BERT \cite{reimers2019sentence} for representation. Our proposed model incorporates such knowledge representation into identifying both explicit and implicit hate speech in investigating the effectiveness of real world knowledge. \par
Overall, this study makes the following contributions: (i) To the best of our knowledge, this work is the first attempt to leverage EL techniques in tackling the problem of implicit hate speech detection. (ii) To evaluate the effectiveness of real world knowledge in both explicit and implicit hate speech detection, where we investigate how incorporating Wikipedia descriptions of linked entities into the model affects performance.

\section{Related Work}

\label{sec:2}
Identifying hate speech has been a topic of immense interest in recent years, and a number of studies have approached this problem in different ways.\par
Early work on hate speech detection has focused on explicitly abusive text using keyword-based methods that rely on lexical features \cite{waseem2016hateful,davidson2017automated}, while more recent studies have highlighted the linguistic nuance and diversity of the implicit hate expressions, which includes stereotypes \cite{sap2019social}, indirect sarcasm, humor, and metaphor \cite{founta2018large} that cannot be captured by keyword-based systems. Implicit hate expressions are no less harmful than explicit ones and make up a large portion of false negatives errors \cite{basile2019semeval,mozafari2020hate}. Systems that rely on explicit hateful lexicon or phrases are unable to capture underlying hateful intent like humans. Up until now, predicting implicit hate or abuse remains a major challenge for machine systems. Existing solutions for identifying implicit cases of hate speech involve taking context into account. For example, \citet{gao2017detecting} included original news articles as the context of the hateful comments. Other studies have built datasets with ``implicit'' labels or annotations \cite{caselli2020feel,elsherief2021latent,sap2019social}. This is crucial not only for evaluation but also for training, as systems that are not trained on implicit hate would not go beyond explicit features and are thus far from being applicable in the real world as a moderation tool. \par
Recently, an emerging line of research has started to explore the idea of incorporating real world knowledge in a related task, sarcasm detection, but not for hate speech detection task. This line of research \cite{chowdhury2021does,li2021sarcasm} hypothesizes that infusing real world knowledge such as commonsense knowledge in sarcasm detection ideally should make the learning model easier to detect sarcasm where it is not apparent from the text. \citet{li2021sarcasm} proposed a novel architecture to integrate knowledge into learning model. For knowledge representation, they applied the pre-trained COMET model (\textbf{COM}mons\textbf{E}nse \textbf{T}ransformers, \citet{bosselut2019comet}) to generate relevant commonsense knowledge from sarcastic instances and use it as input to the proposed model in investigating how commonsense knowledge influences performance. Similarly, \citet{chowdhury2021does} leveraged COMET to infuse commonsense knowledge in their graph convolution-based model, in which a graph is formed with edges between the input sentence and COMET sequences. The node representations of the graph are then passed through a fully-connected neural network to generate the output. \par
The results of the effectiveness of commonsense knowledge are still inconclusive. \citet{li2021sarcasm} found that integrating commonsense knowledge information contributes to sarcasm detection, yet it only plays a supporting role as models using only knowledge information do not perform satisfactorily. Interestingly, \citet{chowdhury2021does} found an opposite result on the role of commonsense knowledge in sarcasm detection, showing that COMET infused model performs at par with the baseline. In many cases, the model is more reliant on the input sentence and less on the COMET sequences for making the prediction. Their in-depth error analysis shows that commonsense is most effective in identifying sarcasm with polarity contrast but fails to explain non-sarcastic samples or other types of sarcasm effectively. The study suggests that exploring the utility of other forms of external knowledge such as factual world knowledge for sarcasm detection would be a promising line of inquiry. Inspired by previous research, the present study assumes that real world knowledge would be beneficial to implicit hate speech detection in that such knowledge allows a fully understanding of the background knowledge required for deciphering the hateful intent of the text.

\section{Methods}
\subsection{Dataset}

In this work, our proposed model is evaluated on the \textit{Latent Hatred Dataset} \cite{elsherief2021latent}. The dataset used in this study contains 12,143 tweets from the most prominent extremist groups in the United States, where 5,791 of these tweets are implicit hate speech and 567 of them are explicit hate speech. Implicit hate tweets are categorized into six classes using the taxonomy shown in Table~\ref{tb:1}.

\begin{table*}[htbp]
\centering
\resizebox{\textwidth}{!}{%
\begin{tabular}{|p{4cm}|p{6cm}|p{6cm}|}
\toprule
\textbf{Class (distribution)} & \textbf{Explanation} & \textbf{Example} \\ \midrule
Grievance (24.2\%) & Frustration over a minority group's perceived privilege & \textit{Black lives matter and white lives don’t? Sounds racist.}\\\hline
Incitement (20\%) & Implicitly promoting known hate groups and ideologies (e.g., by flaunting in-group power) & \textit{Hitler was Germany –Germans shall rise again!} \\\hline
Inferiority (13.6\%) & Implying some group or person is of lesser value than another & \textit{It’s not a coincidence the best places to live are majority white.} \\\hline
Irony (12.6\%) & Using sarcasm, humor, and satire to demean someone & \textit{Horrors... Disney will be forced into hiring Americans.} (Discredit Disney for allegedly hiring only non-whites) \\\hline
Stereotypes (17.9\%)& Associating a group with negative attribute using euphemisms, circumlocution, or metaphorical language & \textit{Can someone tell the black people in Chicago to stop killing one another before it becomes Detroit?} \\\hline
Threats (10.5\%) & Making an indirect commitment to attack someone's body, well-being, reputation, liberty, etc. Focus on threats related to implicit violation of rights and freedoms, removal of opportunities, and more subtle forms of intimidation & \textit{All immigration of non-whites should be ended.} (Non-white immigrants should stay in their country: subtle forms of intimidation) \\\bottomrule
\end{tabular}%
}
\caption{Implicit hate classes and examples in \textit{Latent Hatred Dataset}.}
\label{tb:1}
\end{table*}
Each of the 5,791 implicit hate tweets also has free-text annotations for the target demographic group and an implied statement to describe the underlying message. Implied statements are generated by human annotators with the format \textit{$\langle target \rangle$ \{do, are, commit\}  $\langle predicate \rangle$}, where \textit{$\langle target \rangle$} might be phrases like \textit{immigrants, minorities}. For example, the implicit hate tweet \textit{this selfie is so white, i love it.} has the implied statement ``Minorities are less than whites''. 
\subsection{Models}
In this paper, two classification tasks are conducted. (1) a binary classification task on distinguishing hate speech from non-hate speech, and (2) a 6-way classification task on categorizing implicit hate speech classes (see Table~\ref{tb:1}). \par
For both tasks, a Multi-layer Perceptron (MLP) model with Sentence BERT \cite{reimers2019sentence} embeddings is used. First, we pre-processed all tweets and background knowledge descriptions (remove stop words and reserved words such as RT, FAV, via, etc. while keeping all the hashtags and links that may contain useful messages). Next, they are concatenated before being encoded with Sentence BERT using the pre-trained model \texttt{bert-base-nli-mean-tokens}, where it maps tweets and background knowledge descriptions to a 768-dimensional dense vector space. The MLP model is evaluated with two feature sets: Sentence BERT encoded textual embedding alone (baseline, tweet only) and the combination of textual embedding and background knowledge (baseline+BK). 
\subsection{Background Knowledge Extraction and Representation} \label{sec:3.3}
To incorporate background knowledge, entity linking is applied to associate mentions with their referent entities. First, mentions in each tweet are identified and linked to entities in the KB using Radboud Entity Linker (REL, \citet{van2020rel}), an end-to-end entity linker that identifies mentions of specific entities in text and links them to pertinent Wikipedia page titles. REL is chosen because it has state-of-the-art performance and is trained on a recent Wikipedia dump (2019-07). It provides a web API \footnote{https://rel.cs.ru.nl/api} in which given an input text it returns a list of mentions with the linked entities and the confidence score of mention detection and entity disambiguation. In order to refine the entity linking results (see Table~\ref{tb:2} for an example), we tested different thresholds of confidence score and decided to remove entities with a low confidence score of mention detection (MD score) ($<$0.4) and a low confidence score of entity disambiguation (ED score) ($<$0.2). We find that this refinement strategy helps us strike a balance between precision and recall in that it matches as many mentions as possible (retain mentions that have an MD score $>$ 0.4) while maintaining the accuracy of the result at the same time (remove entities that have an ED score $<$ 0.2). \par
After retrieving all the Wikipedia page titles of the entities in the input text, we use \texttt{Wikipedia} API \footnote{https://pypi.org/project/wikipedia/} to extract the summary of the corresponding Wikipedia page (referred to as \textit{entity abstract} or \textit{Wikipedia description} afterward). To keep the entity abstract from being too long, we print only two sentences of each abstract by setting the \texttt{sentences} argument to \texttt{2}. An example of entity abstract is as follows: \textit{David Ernest Duke (born July 1, 1950) is an American white supremacist, antisemitic conspiracy theorist, far-right politician, convicted felon, and former Grand Wizard of the Knights of the Ku Klux Klan. From 1989 to 1992, he was a member of the Louisiana House of Representatives for the Republican Party.}\par
Finally, the entity abstract for each tweet is concatenated with the tweet and encoded with 768-dimensional Sentence BERT embedding using \texttt{bert-base-nli-mean-tokens} model.

\begin{table*}[htb]
\centering
\resizebox{\textwidth}{!}{

\begin{tabular}{|p{6cm}|p{9cm}|p{9cm}|} 
 \toprule
 \textbf{Tweet} & \textbf{Before Refinement Strategy} & \textbf{After Refinement Strategy (Removing entities with a MD score $<$0.4 and a ED score $<$0.2)} \\ 
 \midrule
 tune in today's jmt for my interview w /robert spencer on ``the complete infidel's guide to iran!" & tune in today's [jmt]\textsubscript{Jedi\_Mind\_Tricks (an American hip hop group)} (ED score: 0.78, MD score: 0.36) for my interview w /[robert spencer] \textsubscript{Robert B. Spencer (American author and blogger, opponent of Islam)} (ED score: 0.38, MD score: 0.96) on ``the complete infidel's guide to [iran] \textsubscript{Iran}'' (ED score: 0.51, MD score: 0.99) & tune in today's jmt for my interview w /[robert spencer] \textsubscript{Robert B. Spencer (American author and blogger, opponent of Islam)} (ED score: 0.38, MD score: 0.96) on ``the complete infidel's guide to [iran] \textsubscript{Iran}'' (ED score: 0.51, MD score: 0.99)\\\bottomrule
\end{tabular}
}

\caption{An Entity linking example from our dataset.}
\label{tb:2}

\end{table*}

\section{Experiments}\label{sec:4}
\subsection{Experimental Setup}
For each of the two classification tasks, the model is trained and evaluated on two feature sets, which are baseline feature set (tweet text only) with and without Wikipedia descriptions. For both binary and 6-way classification task, a MLP is implemented in \texttt{sklearn} with three hidden layers of dimension 512, learning rate \texttt{0.001} and the number of epochs \texttt{500}. The optimizer is set to \texttt{Adam}. 
\subsection{Classification Results}\label{sec:4.2}
In explicit hate speech classification shown in Table~\ref{tb:3}, the background knowledge provided by Wikipedia significantly improves the model by increasing 10\% in precision, recall, and F1 score after incorporating background knowledge into the model. While the MLP model with baseline feature set achieves a competitive result with 65\% on F1 score, the background knowledge incorporated model achieves better scores (75\%) than the one with baseline feature set, demonstrating that real world knowledge is helpful for capturing real hate speech. \par
For additional comparisons \footnote{We notice that there are works \cite{pal2022combating} that also improve performance on the \textit{Latent Hatred Dataset}, but we compare our results against the results from \citet{elsherief2021latent}, which serves as a benchmark for modeling implicit hate speech using knowledge-based features as well.}, our background knowledge incorporated model achieves a significantly better precision score (75\% vs. 68\%) than the Wikidata Knowledge Graph \cite{vrandevcic2014wikidata} infused model proposed in \citet{elsherief2021latent}, which was trained on the same \textit{Latent Hatred Dataset}. The remarkably higher precision score suggests that Wikipedia description of linked entities is doing a better job in preventing false positives than Wikidata Knowledge Graph method; however, a more detailed analysis comparing the effectiveness of different external knowledge (e.g., knowledge graphs, commonsense knowledge) for hate speech detection is needed. 
\begin{table*}[t]
\begin{center}
\resizebox{\linewidth}{!}{%
\begin{tabular}{lrrrrp{6cm}l} 
 \toprule
 \textbf{Models} & \textbf{P} & \textbf{R} & \textbf{F} & \textbf{Acc} \\ 
 \midrule
Majority baseline & \multicolumn{4}{l}{$\SI{52}{\percent} \pm \SI{1.3}{\percent}$} \\
MLP (baseline) & $\SI{65}{\percent} \pm \SI{1.5}{\percent}$ & $\SI{65}{\percent} \pm \SI{1.5}{\percent}$ & $\SI{65}{\percent} \pm \SI{1.5}{\percent}$ & $\SI{65}{\percent} \pm \SI{1.5}{\percent}$ \\
MLP (baseline+BK) & $\SI{75}{\percent} \pm \SI{1.4}{\percent}$ & $\SI{75}{\percent} \pm \SI{1.4}{\percent}$ & $\SI{75}{\percent} \pm \SI{1.4}{\percent}$ & $\SI{75}{\percent} \pm \SI{1.4}{\percent}$\\
Knowledge infused model \cite{elsherief2021latent} & 68\% & 72\% & 70\% & \textbf{77\%}\\\bottomrule
\end{tabular}%
}
\caption{Classification performance on explicit hate speech classification. Performance metrics are all macro average scores. Majority baseline always returns the positive (hate) label. \textbf{Bold} face indicates best performance.}
\label{tb:3}
\end{center}
\end{table*}

As shown in Table~\ref{tb:3}, real world knowledge enhances the correct detection of explicit hate speech. However, Table~\ref{tb:4} shows that integrating real world knowledge does not seem to improve the model, and even hurts model's performance in implicit hate speech type classification. Significant degradation in precision, recall, and F1 score are observed (e.g., recall drops 12\%), which suggests that knowledge about the involved entities is not sufficient for predicting implicit hate speech types. \par
Table~\ref{tb:5} further shows that among the six implicit hate classes, \textit{irony} is the hardest for the model to detect, which aligns with the result found in \citet{elsherief2021latent}. This is reasonable, as irony normally requires further understanding beyond knowledge about the involved entities (e.g., semantic inference or pragmatic understanding). Therefore, our background knowledge incorporated model fails at capturing this type of implicit hate. On the other hand, \textit{white grievance} is the easiest to detect for our model. A detailed examination of our data shows that compared to other types of implicit hate posts, \textit{white grievance} tweets in our dataset contain relatively more explicit hate triggers (e.g., rally, KKK), which is found to be useful for the model. Further discussion on explaining our model's predictions on implicit hate could be found in Section~ \ref{sec:5}. 
 
\begin{table*}[t]
\begin{center}
\begin{tabular}{lrrrrp{6cm}l} 
 \toprule
 \textbf{Feature sets} & \textbf{P} & \textbf{R} & \textbf{F} & \textbf{Acc}  \\ 
 \midrule
 Dummy Classifier & $\SI{19}{\percent} \pm \SI{1.0}{\percent}$ \\
 Baseline (tweet only) & $\SI{52}{\percent} \pm \SI{1.3}{\percent}$ & $\SI{52}{\percent} \pm \SI{1.3}{\percent}$ & $\SI{52}{\percent} \pm \SI{1.3}{\percent}$ & $\bm{54\%} \pm \SI{1.3}{\percent}$  \\
 Baseline + BK & $\SI{42}{\percent} \pm \SI{1.3}{\percent}$ & $\SI{40}{\percent} \pm \SI{1.3}{\percent}$ & $\SI{41}{\percent} \pm \SI{1.3}{\percent}$ & $\SI{44}{\percent} \pm \SI{1.3}{\percent}$ \\\bottomrule
\end{tabular}
\caption{Classification performance on 6-way implicit hate speech classification. Performance metrics are all macro average scores. Dummy classifier generates random predictions by respecting the training set class distribution. \textbf{Bold} face indicates best performance.}
\label{tb:4}
\end{center}
\end{table*}

\begin{table*}[t]
\begin{center}
\resizebox{\linewidth}{!}{%
\begin{tabular}{r r r r r r r} 
 \toprule
 \textbf{Feature sets} & \textbf{incitement} & \textbf{inferiority} & \textbf{irony} & \textbf{stereotypical} & \textbf{threatening} & \textbf{white grievance}  \\ 
 \midrule
MLP (baseline) & 52\% & 51\% & \textit{33}\% & 55\% & 56\% & \textbf{62\%} \\
MLP (baseline+BK) & 45\% & 40\% & \textit{20}\% & 44\% & 43\% & 52\%\\\bottomrule
\end{tabular}%
}
\caption{Classification performance on 6-way implicit hate speech classification. Performance metrics are all F1 scores. \textit{Italics} indicates the worst performance. \textbf{Bold} face indicates best performance.}
\label{tb:5}
\end{center}
\end{table*}

\section{Analysis} \label{sec:5}
This section investigates whether the model is reliant on Wikipedia descriptions while making decisions. We leverage LIME (locally interpretable model-agnostic  explanations) algorithms \cite{ribeiro2016lime} to explain our model's predictions on both explicit and implicit hate statements through random examples picked from our dataset.

\subsection{Efficacy of World Knowledge}\label{sec:5.1}
Table~\ref{tb:6} shows cases where Wikipedia knowledge is helpful. For each example, the colored text spans represent the words highly-weighted by the model. We find that Wikipedia knowledge is particularly useful when hatefulness in a tweet is conveyed through certain hate ``triggers''. These triggers by themselves are not toxic but are relevant to the hatefulness in a tweet. Since implicit hate does not contain explicit hate lexicon or phrases, the model rather relies on these triggers to help them make the right predictions. As shown in Example C of Table~\ref{tb:6}, the Wikipedia description of \textit{Charlottesville} helps the model correctly predicts the tweet as an incitement tweet by relying on entity triggers such as \textit{rally}. The word by itself is not a hate lexicon but indicates a high probability of a tweet that incites violence. Similarly, the Wikipedia description of \textit{David Duke} in Example B of Table~\ref{tb:6} is helpful for the model in that it explains \textit{David Duke} is the former head of \textit{Ku Klux Klan}, which by itself does not convey toxicity but is indicative of a \textit{white grievance} tweet. \par
Table~\ref{tb:6} further shows that although entity triggers provided by Wikipedia description contribute to the detection of hate speech to a certain extent, it only plays a supporting role. Some of the words in the tweet are already a strong signal of the hatefulness of the tweet, as shown in the highlighted words in the table. For instance, in Example B of Table~\ref{tb:6}, both the hashtag \textit{\#makeamericagreatagain} and \textit{\#votetrump} in the post reveal that the author might be a supporter of Donald Trump, which is said to have a symbiotic relationship with white nationalism, white supremacy, and white power ideologies that correspond to the \textit{white grievance} implicit hate type in our dataset.
\begin{table*}[t]
\centering
\resizebox{\textwidth}{!}{%
\begin{tabular}{|p{1cm}|p{2cm}|p{2cm}|p{13.5cm}|}
\toprule
\# & Gold & Prediction & Example \\ \midrule
A & Explicit hate &
  Explicit hate &
  \begin{tabular}[c]{@{}l@{}}\textit{the kkk. how \colorbox{OrangeRed1}{old} does my little boy need to be to join?}\\ \textbf{The Ku Klux Klan (KKK), or simply "the Klan", is the name of three distinct} \\\textbf{past and present movements in the United States that have advocated extremist} \\ \textbf{reactionary currents such as white supremacy, white nationalism, and anti-}\\\textbf{\colorbox{OrangeRed3}{immigration}, historically expressed through \colorbox{OrangeRed3}{terrorism} aimed at groups or}\\ \textbf{individuals whom they opposed.}\end{tabular} \\\hline
B &White grievance &
  White grievance &
  \begin{tabular}[c]{@{}l@{}}\textit{refuses to denounce dr.david duke! \#\colorbox{OrangeRed3}{makeamericagreatagain} \#\colorbox{OrangeRed3}{votetrump}}\\ \textbf{David \colorbox{OrangeRed1}{Ernest} Duke (born July 1, 1950) is an American white nationalist,} \\\textbf{anti-Semitic conspiracy theorist, politician, and former Grand Wizard of the Ku} \\ \textbf{Klux \colorbox{OrangeRed3}{Klan}.}\end{tabular} \\\hline
C & Incitement &
  Incitement &
  \begin{tabular}[c]{@{}l@{}}\textit{\#\colorbox{OrangeRed2}{charlottesville} a \colorbox{OrangeRed2}{day} that will go down in white history.}\\ \textbf{\colorbox{OrangeRed2}{Charlottesville}, a metonym for the \colorbox{OrangeRed3}{Unite} the Right \colorbox{OrangeRed3}{rally}, a white} \\\textbf{supremacist \colorbox{OrangeRed3}{rally} that took place in \colorbox{OrangeRed2}{Charlottesville}, Virginia, from August 11} \\ \textbf{to 12, 2017.}\end{tabular} \\ \bottomrule
\end{tabular}
}
\caption{Examples where Wikipedia description helps the model make the right decision. \colorbox{OrangeRed3}{Red} color indicates the features with highest coefficients for the model, darker colors indicate more polarity. The tweet is in \textit{italics}. Wikipedia description is indicated in \textbf{bold}.}
\label{tb:6}
\end{table*}

\subsection{Error Analysis} \label{sec:5.2}
To further understand the role of world knowledge in identifying hatefulness, we randomly pick out incorrect predictions made by our model and manually correct some of the entity linking errors to see if this ``post-processing'' helps avert classification errors. \par
As shown in Example A of Table~\ref{tb:7}, our entity linker misses the mention \textit{bernie bros}. Instead, \textit{white males} in the tweet is identified and linked to Wikipedia. The Wikipedia description of \textit{white males} does not add value to the detection of hatefulness. Words with the highest coefficients such as \textit{skin} and \textit{African} are neutral and are not associated with the hateful content of the tweet. After post-processing the entity linking result, our model correctly predicts the tweet as a hate post by leveraging the world knowledge provided by Wikipedia. The Wikipedia description explains that \textit{bernie bros} is a pejorative term used to describe Bernie Sanders supporters that have recently received criticism for crude and sexist attacks against rival Hillary Clinton. A similar example is shown in Example B of Table~\ref{tb:7}, where the term \textit{Charlottesville} is used to refer to the white supremacist rally that took place in Charlottesville, Virginia rather than as a city's name. Before post-processing, our model incorrectly predicts the tweet as a \textit{threatening} post based on the words used in the post (\textit{worry, fast}). By contrast, our model accurately detects the implicitly hateful message conveyed in the tweet after  \textit{Charlottesville} is being identified in its correct sense.
\begin{table*}[htbp]
\centering
\resizebox{\textwidth}{!}{%
\begin{tabular}{|p{0.5cm}|p{2cm}|p{3cm}|p{16.5cm}|l}
\toprule
\# &
  Gold &
  Model predictions before and after post-processing &
  Example &
   \\ \midrule
\multirow{2}{*}{A} &
  \multirow{2}{*}{Explicit hate} &
  Non-hate &
  \begin{tabular}[c]{@{}l@{}}\textit{bernie bros are entitled white males \#maleprivileage \#culturalmarxism \#\colorbox{OrangeRed1}{feminismiscamcer} \#\colorbox{OrangeRed2}{democrats}}\\ \textbf{White is a racialized classification of people and a \colorbox{OrangeRed1}{skin} color specifier, generally used for people of} \\ \textbf{European origin; although the definition can vary depending on context, nationality, and point of}\\ \textbf{view. In the United States, this term has at times been expanded to encompass persons of Mexican,}\\ \textbf{South Asian, West Asian, East Asian, and North \colorbox{OrangeRed1}{African} descent, persons who are often considered}\\\textbf{ ``non-White" in other contexts in the United States.}\end{tabular} &
   \\\cmidrule{3-4}
 &
   &
  Explicit hate &
  \begin{tabular}[c]{@{}l@{}}\textit{bernie bros are entitled white males \#maleprivileage \#culturalmarxism \#\colorbox{OrangeRed2}{feminismiscamcer} \#\colorbox{OrangeRed2}{democrats}}\\ \textbf{"Bernie Bro", or sometimes "Berniebro", is a \colorbox{OrangeRed3}{pejorative} term for a person (particularly one who} \\ \textbf{is young, male and wealthy) who} \textbf{supports American politician and junior United States senator} \\ \textbf{Bernie Sanders via Internet culture and social media by ``mainstream" media outlets and later} \\ \textbf{Hillary Clinton supporters. Recently, Bernie Bros have received criticism for crude and sexist}\\ \textbf{\colorbox{OrangeRed3}{attacks} against rival Democratic} \textbf{nominee Hillary Clinton and her second presidential campaign}\\ \textbf{in 2016, with her first attempt being in 2008.}\end{tabular} &
   \\\cmidrule(r){1-4}
\multirow{2}{*}{B} &
  \multirow{2}{*}{Incitement} &
  Threatening &
  \begin{tabular}[c]{@{}l@{}}\textit{don't \colorbox{OrangeRed2}{worry} charlottesville was just the beginning. we're growing extremely \colorbox{OrangeRed2}{fast}.}\\ \textbf{Charlottesville, colloquially C'ville, Hoo-Ville, and formally the City of Charlottesville, is an} \\ \textbf{independent city in the Commonwealth of Virginia. As of 2015, the \colorbox{OrangeRed1}{population} was 46,597.}\end{tabular} \\\cmidrule{3-4}
 &
   &
  Incitement &
  \begin{tabular}[c]{@{}l@{}}\textit{don't \colorbox{OrangeRed2}{worry} charlottesville was just the beginning. we're growing extremely \colorbox{OrangeRed2}{fast}.}\\ \textbf{Charlottesville, a metonym for the \colorbox{OrangeRed3}{Unite} the Right \colorbox{OrangeRed3}{rally}, a white supremacist \colorbox{OrangeRed3}{rally} that took} \\ \textbf{place in Charlottesville, Virginia,} \textbf{from August 11 to 12, 2017.}\end{tabular} &
   \\ \bottomrule
\end{tabular}%
}
\caption{Ground truth labels along with predictions made by the model before and after post-processing. \colorbox{OrangeRed3}{Red} color indicates the features with the highest coefficients for the model, darker colors indicate more polarity. The tweet is in \textit{italics}. Wikipedia description is indicated in \textbf{bold}.}
\label{tb:7}
\end{table*}

\subsection{Challenges in Implicit Hate Speech Detection}
\label{sec:5.3}
Section~\ref{sec:4.2} shows that Wikipedia knowledge does not add value and sometimes even hurts the implicit hate speech type classification. To further understand the role of Wikipedia knowledge in implicit hate type classification, we identify two challenges our model struggles with in implicit hate speech classification. One of the most common challenges for our model is shown in \ref{ex3}, where hatred is not apparent from the text but requires a pragmatic understanding of the text. 

\ex. \textit{\colorbox{OrangeRed3}{white} \colorbox{OrangeRed2}{privilege} conference this week. \#nationalsocialism} \textbf{\colorbox{OrangeRed3}{white} \colorbox{OrangeRed2}{privilege}, or \colorbox{OrangeRed3}{white} skin \colorbox{OrangeRed2}{privilege}, is the societal \colorbox{OrangeRed2}{privilege} that benefits \colorbox{OrangeRed3}{white} people over non-white people in some societies, particularly if they are otherwise under the same social, political, or economic circumstances. With roots in European colonialism and imperialism, and the Atlantic slave trade, \colorbox{OrangeRed3}{white} privilege has developed in circumstances that have broadly sought to protect \colorbox{OrangeRed3}{white} \colorbox{OrangeRed2}{racial} privileges, various national citizenships, and other rights or special benefits.} (Gold label: Incitement, Prediction: White grievance)\label{ex3}

\ex. \textit{\colorbox{OrangeRed2}{atlanta} hawks poor attendance is blamed on \colorbox{OrangeRed3}{black} \colorbox{OrangeRed2}{crowds} making southern whites uncomfortable.} \textbf{The \colorbox{OrangeRed2}{Atlanta} Hawks are a professional basketball team based in \colorbox{OrangeRed2}{Atlanta}, Georgia. The Hawks compete in the National Basketball Association (NBA) as a member team of the league's Eastern Conference Southeast Division. The Hawks play their home games at Philips Arena. } (Gold: Incitement, Prediction: Stereotypical )\label{ex4}

The post in \ref{ex3} is an \textit{incitement} post because it implicitly elevates white privilege ideology by promoting an upcoming event on white privilege. However, the tweet is predicted as a \textit{white grievance} post because of the recurrence of the term \textit{white privilege} in the post and Wikipedia description. Here we can see that world knowledge distracts the model. Additionally, our model struggles with this type of indirect hate because the implied hateful message is not apparent from the text. The model has to understand the pragmatic implicature (White people are privileged) that the post suggests or implies in order to decipher the hidden hateful intent of the tweet.\par
\ref{ex4} demonstrates another common challenge our model faces. The description of the basketball team does not help understand the underlying hatefulness of the tweet, which implies that white people don't go to watch the basketball game because black players would make white people uncomfortable. That said, the Wikipedia description here does not hurt the model as well because the model still predicts the tweet as \textit{stereotypical} without the description. The underlying problem here is that Wikipedia descriptions and tweet text seem to be unrelated to each other. This domain discrepancy between text and knowledge suggests that simple concatenation is not enough, but a more sophisticated structure that can capture the information flow between text and knowledge representation is needed for implicit hate speech type classification.

\section{Conclusion}
This paper has proposed the idea of integrating real world knowledge into the task of hate speech detection. Experimental results show that real world knowledge is helpful, especially in cases where entity triggers (e.g., rally, KKK) are present in the tweet. However, our analysis also shows that this knowledge fails to predict implicit hate speech types as Wikipedia knowledge does not add value and sometimes even hurts the classification, suggesting that a more sophisticated model that enables understanding beyond knowledge about the involved entities is required for implicit hate speech type classification. To mitigate these challenges, works on model architecture that enable information flow between the representations of the tweet and Wikipedia knowledge is a reasonable next step. Additionally, exploring the possibility of combining different kinds of external knowledge, for example, combining commonsense knowledge \cite{chowdhury2021does} in modeling implicit hate speech would also be a promising line of inquiry. To further understand the subtleties of hate speech, deciphering models for coded language or indirect language (e.g., metaphor, irony) in hate speech expression would be beneficial.

\section*{Ethics Statement}
With the exponential growth of offensive language online, a myriad of machine learning models has been proposed. However, a major limitation of many existing hate speech detection models is that they focused on capturing explicit or overt hate speech, failing to detect implicit hateful expressions that are no less harmful than explicit ones. Our experiment in this study suggests that entity triggers (e.g., rally, KKK) are helpful in detecting hatefulness that is not apparent from the text. This could ideally help improve model accuracy in identifying implicit hate speech, preventing targeted communities from experiencing increased harm online. Furthermore, we show that knowledge about entities may help reduce false positives in explicit hate speech. This is important, as deep learning models nowadays still suffer from false positive predictions \cite{markov-daelemans-2021-improving}. To this end, minimizing false positives is pivotal, as models that are not robust enough would be far from being applicable in the real world as a moderation tool, and using such a non-robust model would further lead to over-blocking or removal of harmless social media content that does not violate community guidelines. 


\bibliography{anthology,custom}

\begin{thebibliography}{20}
\expandafter\ifx\csname natexlab\endcsname\relax\def\natexlab#1{#1}\fi

\bibitem[{Basile et~al.(2019)Basile, Bosco, Fersini, Debora, Patti, Pardo,
  Rosso, Sanguinetti et~al.}]{basile2019semeval}
Valerio Basile, Cristina Bosco, Elisabetta Fersini, Nozza Debora, Viviana
  Patti, Francisco Manuel~Rangel Pardo, Paolo Rosso, Manuela Sanguinetti,
  et~al. 2019.
\newblock Semeval-2019 task 5: Multilingual detection of hate speech against
  immigrants and women in twitter.
\newblock In \emph{13th International Workshop on Semantic Evaluation}, pages
  54--63. Association for Computational Linguistics.

\bibitem[{Bosselut et~al.(2019)Bosselut, Rashkin, Sap, Malaviya, Celikyilmaz,
  and Choi}]{bosselut2019comet}
Antoine Bosselut, Hannah Rashkin, Maarten Sap, Chaitanya Malaviya, Asli
  Celikyilmaz, and Yejin Choi. 2019.
\newblock Comet: Commonsense transformers for automatic knowledge graph
  construction.
\newblock \emph{arXiv preprint arXiv:1906.05317}.

\bibitem[{Caselli et~al.(2020)Caselli, Basile, Jelena, Inga, Michael
  et~al.}]{caselli2020feel}
Tommaso Caselli, Valerio Basile, Mitrovi{\'c} Jelena, Kartoziya Inga, Granitzer
  Michael, et~al. 2020.
\newblock I feel offended, don’t be abusive! implicit/explicit messages in
  offensive and abusive language.
\newblock In \emph{Language Resources and Evaluation Conference}, pages
  6193--6202. The European Language Resources Associatio.

\bibitem[{Chowdhury and Chaturvedi(2021)}]{chowdhury2021does}
Somnath Basu~Roy Chowdhury and Snigdha Chaturvedi. 2021.
\newblock Does commonsense help in detecting sarcasm?
\newblock \emph{arXiv preprint arXiv:2109.08588}.

\bibitem[{Davidson et~al.(2017)Davidson, Warmsley, Macy, and
  Weber}]{davidson2017automated}
Thomas Davidson, Dana Warmsley, Michael Macy, and Ingmar Weber. 2017.
\newblock Automated hate speech detection and the problem of offensive
  language.
\newblock In \emph{Proceedings of the International AAAI Conference on Web and
  Social Media}, volume~11, pages 512--515.

\bibitem[{ElSherief et~al.(2021)ElSherief, Ziems, Muchlinski, Anupindi,
  Seybolt, De~Choudhury, and Yang}]{elsherief2021latent}
Mai ElSherief, Caleb Ziems, David Muchlinski, Vaishnavi Anupindi, Jordyn
  Seybolt, Munmun De~Choudhury, and Diyi Yang. 2021.
\newblock Latent hatred: A benchmark for understanding implicit hate speech.
\newblock \emph{arXiv preprint arXiv:2109.05322}.

\bibitem[{Founta et~al.(2018)Founta, Djouvas, Chatzakou, Leontiadis, Blackburn,
  Stringhini, Vakali, Sirivianos, and Kourtellis}]{founta2018large}
Antigoni~Maria Founta, Constantinos Djouvas, Despoina Chatzakou, Ilias
  Leontiadis, Jeremy Blackburn, Gianluca Stringhini, Athena Vakali, Michael
  Sirivianos, and Nicolas Kourtellis. 2018.
\newblock Large scale crowdsourcing and characterization of twitter abusive
  behavior.
\newblock In \emph{Twelfth International AAAI Conference on Web and Social
  Media}.

\bibitem[{Gao and Huang(2017)}]{gao2017detecting}
Lei Gao and Ruihong Huang. 2017.
\newblock Detecting online hate speech using context aware models.
\newblock \emph{arXiv preprint arXiv:1710.07395}.

\bibitem[{Kiritchenko et~al.(2021)Kiritchenko, Nejadgholi, and
  Fraser}]{kiritchenko2021confronting}
Svetlana Kiritchenko, Isar Nejadgholi, and Kathleen~C Fraser. 2021.
\newblock Confronting abusive language online: A survey from the ethical and
  human rights perspective.
\newblock \emph{Journal of Artificial Intelligence Research}, 71:431--478.

\bibitem[{Li et~al.(2021)Li, Pan, Lin, Fu, and Wang}]{li2021sarcasm}
Jiangnan Li, Hongliang Pan, Zheng Lin, Peng Fu, and Weiping Wang. 2021.
\newblock Sarcasm detection with commonsense knowledge.
\newblock \emph{IEEE/ACM Transactions on Audio, Speech, and Language
  Processing}, 29:3192--3201.

\bibitem[{Markov and Daelemans(2021)}]{markov-daelemans-2021-improving}
Ilia Markov and Walter Daelemans. 2021.
\newblock \href {https://doi.org/10.18653/v1/2021.nlp4if-1.3} {Improving
  cross-domain hate speech detection by reducing the false positive rate}.
\newblock In \emph{Proceedings of the Fourth Workshop on NLP for Internet
  Freedom: Censorship, Disinformation, and Propaganda}, pages 17--22, Online.
  Association for Computational Linguistics.

\bibitem[{Mozafari et~al.(2020)Mozafari, Farahbakhsh, and
  Crespi}]{mozafari2020hate}
M~Mozafari, R~Farahbakhsh, and N~Crespi. 2020.
\newblock Hate speech detection and racial bias mitigation in social media
  based on bert model.
\newblock \emph{PLoS ONE}, 15(8):e0237861.

\bibitem[{Pal et~al.(2022)Pal, Chaudhari, and Sharma}]{pal2022combating}
Debaditya Pal, Kaustubh Chaudhari, and Harsh Sharma. 2022.
\newblock Combating high variance in data-scarce implicit hate speech
  classification.
\newblock \emph{arXiv preprint arXiv:2208.13595}.

\bibitem[{Reimers and Gurevych(2019)}]{reimers2019sentence}
Nils Reimers and Iryna Gurevych. 2019.
\newblock Sentence-bert: Sentence embeddings using siamese bert-networks.
\newblock \emph{arXiv preprint arXiv:1908.10084}.

\bibitem[{Ribeiro et~al.(2016)Ribeiro, Singh, and Guestrin}]{ribeiro2016lime}
Marco~Tulio Ribeiro, Sameer Singh, and Carlos Guestrin. 2016.
\newblock " why should i trust you?" explaining the predictions of any
  classifier.
\newblock In \emph{Proceedings of the 22nd ACM SIGKDD international conference
  on knowledge discovery and data mining}, pages 1135--1144.

\bibitem[{Sap et~al.(2019)Sap, Gabriel, Qin, Jurafsky, Smith, and
  Choi}]{sap2019social}
Maarten Sap, Saadia Gabriel, Lianhui Qin, Dan Jurafsky, Noah~A Smith, and Yejin
  Choi. 2019.
\newblock Social bias frames: Reasoning about social and power implications of
  language.
\newblock \emph{arXiv preprint arXiv:1911.03891}.

\bibitem[{van Hulst et~al.(2020)van Hulst, Hasibi, Dercksen, Balog, and
  de~Vries}]{van2020rel}
Johannes~M van Hulst, Faegheh Hasibi, Koen Dercksen, Krisztian Balog, and
  Arjen~P de~Vries. 2020.
\newblock Rel: An entity linker standing on the shoulders of giants.
\newblock In \emph{Proceedings of the 43rd International ACM SIGIR Conference
  on Research and Development in Information Retrieval}, pages 2197--2200.

\bibitem[{Vrande{\v{c}}i{\'c} and Kr{\"o}tzsch(2014)}]{vrandevcic2014wikidata}
Denny Vrande{\v{c}}i{\'c} and Markus Kr{\"o}tzsch. 2014.
\newblock Wikidata: a free collaborative knowledgebase.
\newblock \emph{Communications of the ACM}, 57(10):78--85.

\bibitem[{Waseem et~al.(2017)Waseem, Davidson, Warmsley, and
  Weber}]{waseem2017understanding}
Zeerak Waseem, Thomas Davidson, Dana Warmsley, and Ingmar Weber. 2017.
\newblock Understanding abuse: A typology of abusive language detection
  subtasks.
\newblock \emph{arXiv preprint arXiv:1705.09899}.

\bibitem[{Waseem and Hovy(2016)}]{waseem2016hateful}
Zeerak Waseem and Dirk Hovy. 2016.
\newblock Hateful symbols or hateful people? predictive features for hate
  speech detection on twitter.
\newblock In \emph{Proceedings of the NAACL student research workshop}, pages
  88--93.

\end{thebibliography}
\bibliographystyle{acl_natbib}




\end{document}